\documentclass[runningheads]{llncs}
\usepackage[T1]{fontenc}
\usepackage{amsmath,amssymb}
\usepackage{bm}
\usepackage{booktabs}
\usepackage{caption}
\usepackage{enumitem}
\usepackage{graphicx}
\usepackage{multirow}
\usepackage{hyperref}
\usepackage{placeins}
\usepackage{pifont}     
\usepackage{subcaption}
\usepackage{verbatim}
\usepackage{xcolor}
\usepackage{tabularx}
\usepackage{float}
\usepackage{threeparttable}
\usepackage{multirow}
\usepackage{arydshln}
\usepackage{url}
\usepackage{newtxtext}
\usepackage{newtxmath}
\usepackage{textcomp}

\usepackage{comment}

\begin{document}

\title{ADP-DiT: Text-Guided Diffusion Transformer for Brain Image Generation in Alzheimer’s Disease Progression}

\titlerunning{ADP-DiT: Text-Guided Diffusion Transformer for Alzheimer’s Progression}

\makeatletter
\renewcommand{\thefootnote}{\fnsymbol{footnote}}
\renewcommand*\@fnsymbol[1]{\ifcase #1 \or * \or † \else \@ctrerr \fi}
\makeatother

\author{
Juneyong Lee\inst{1}\thanks{These authors contributed equally to this work.}\orcidID{0009-0004-6723-5485} \and
Geonwoo Baek\inst{1}\protect\footnotemark[1]\orcidID{0009-0001-2338-4575} \and
Ikbeom Jang\inst{1}\thanks{Corresponding author.}\orcidID{0000-0002-6901-983X} \and 
for the Alzheimer's Disease Neuroimaging Initiative
}
\authorrunning{J. Lee et al.}
\institute{Department of Computer Science and Engineering,
Hankuk University of Foreign Studies, Seoul, Republic of Korea\\
\email{\{diziyong, phlox3959, ijang\}@hufs.ac.kr}
}

%
%
\maketitle

\begin{abstract} 
Alzheimer’s disease (AD) progresses heterogeneously across individuals, motivating subject-specific synthesis of follow-up magnetic resonance imaging (MRI) to support progression assessment. While Diffusion Transformers (DiT), an emerging transformer-based diffusion model, offer a scalable backbone for image synthesis, longitudinal AD MRI generation with clinically interpretable control over follow-up time and participant metadata remains underexplored.
We present ADP-DiT, an interval-aware, clinically text-conditioned diffusion transformer for longitudinal AD MRI synthesis. ADP-DiT encodes follow-up interval together with multi-domain demographic, diagnostic (CN/MCI/AD), and neuropsychological information as a natural-language prompt, enabling time-specific control beyond coarse diagnostic stages. To inject this conditioning effectively, we use dual text encoders--OpenCLIP for vision–language alignment and T5 for richer clinical-language understanding. Their embeddings are fused into DiT through cross-attention for fine-grained guidance and adaptive layer normalization for global modulation. We further enhance anatomical fidelity by applying rotary positional embeddings to image tokens and performing diffusion in a pretrained SDXL-VAE latent space to enable efficient high-resolution reconstruction.
On 3,321 longitudinal 3T T1-weighted scans from 712 participants (259,038 image slices), ADP-DiT achieves SSIM 0.8739 and PSNR 29.32 dB, improving over a DiT baseline by +0.1087 SSIM and +6.08 dB PSNR while capturing progression-related changes such as ventricular enlargement and shrinking hippocampus. These results suggest that integrating comprehensive, subject-specific clinical conditions with architectures can improve longitudinal AD MRI synthesis.

\end{abstract}

\section{Introduction}

Recent advancements in generative models have led to the rapid emergence of text-conditioned image generation as a powerful paradigm, offering precise control over visual synthesis through natural-language prompts. While these techniques have revolutionized general computer vision and are increasingly being applied to general medical image synthesis, their specific potential for modeling the complex, longitudinal progression of neurodegenerative diseases remains largely unexplored. Although MRI generation itself is an active area of research, studies that specifically simulate Alzheimer's disease (AD) trajectories—by integrating clinical metadata as textual conditions—represent a critical yet under-researched domain. Existing methods primarily focus on static classification or simple image-to-image translation, leaving the capabilities of text-driven progression modeling largely unexplored in the current literature. This scarcity of research limits the ability to visualize subject-specific disease evolution based on comprehensive clinical descriptions. 


In this study, we address this limitation by presenting ADP-DiT, a novel generative framework that synthesizes longitudinal AD progression in brain MRI images using text-guided diffusion transformers. We target this understudied intersection of multimodal conditioning and disease progression modeling, proposing a robust architecture tailored for neuroimaging. Our technical contributions are twofold: First, we refine the Diffusion Transformer (DiT)~\cite{peebles2023dit} architecture by incorporating Rotary Positional Embeddings (RoPE)~\cite{su2024roformer} to improve spatial feature alignment—critical for preserving anatomical fidelity—and utilizing the Variational Autoencoder from the Stable Diffusion XL framework (SDXL-VAE)~\cite{podell2023sdxl} for efficient latent space modeling to enable high-resolution image reconstruction. Second, we employ a multi-text-encoder strategy that synergistically combines OpenCLIP ViT-G/14~\cite{cherti2023reproducible} for high-level semantic alignment with T5-XXL~\cite{raffel2020t5} (Text-to-Text Transfer Transformer) for nuanced understanding of medical language. This dual-encoder approach enables the model to interpret complex clinical metadata and guide the generation process with greater precision. Collectively, this framework facilitates the synthesis of biologically plausible changes in brain morphology over time, providing a robust foundation for longitudinal neuroimaging research and clinical research applications.

\vspace{-0.2cm}
\subsection{Related works}
Text-guided image generation has evolved from CLIP-guided~\cite{radford2021clip} (contrastive language-image pretraining) synthesis methods such as VQGAN-CLIP~\cite{crowson2022vqgan} to diffusion-based text-guided manipulation (DiffusionCLIP)~\cite{kim2022diffusionclip}. In parallel, latent diffusion models (LDMs) such as Stable Diffusion~\cite{rombach2022sd} enable high-fidelity synthesis by operating in a compressed latent space, and variants such as FCDiffusion~\cite{wu2024fcdiffusion} improve controllability for text-guided translation.

Most LDMs rely on CNN U-Net backbones, which can be suboptimal for capturing global anatomical context. DiT~\cite{peebles2023dit} replaces the denoiser with a transformer operating on latent patches, improving long-range dependency modeling. As a deterministic medical baseline, we include SwinUNETR-V2~\cite{hatamizadeh2023swinunetr} to contrast point-estimate prediction with our diffusion-based, progression-conditioned synthesis.

\section{Material and methods}
\subsection{Imaging and participants}
In this study, we utilized 3{,}321 longitudinal MRI scans from $N=712$ participants and accompanying structured demographic and clinical metadata from the Alzheimer’s Disease Neuroimaging Initiative (ADNI)~\cite{petersen2010adni} dataset. All scans were three-dimensional (3D) accelerated sagittal T1-weighted (T1w) acquisitions at 3~T from ADNI~\cite{jack2010} (ADNI-GO/2). The scans were acquired using Siemens MPRAGE, GE IR-SPGR (BRAVO), and Philips IR-TFE (SENSE) protocols, with a voxel size of $1.0 \times 1.0 \times 1.2~\mathrm{mm}^3$. The cohort composition by AD progression group is summarized in Table~\ref{tab:tab1}. The textual metadata includes disease status labels—Cognitively Normal (CN), Mild Cognitive Impairment (MCI), and AD—as well as demographic information (e.g., age) and clinical and cognitive assessments, which are essential for disease progression modeling. These assessments included the Clinical Dementia Rating--Sum of Boxes (CDR-SB), Mini-Mental State Examination (MMSE), Montreal Cognitive Assessment (MoCA), Functional Activities Questionnaire (FAQ), Rey Auditory Verbal Learning Test (RAVLT; immediate, learning, forgetting, and percent forgetting), Logical Memory delayed recall total (LDELTOTAL), Trail Making Test Part B completion time (TRABSCOR), and Alzheimer’s Disease Assessment Scale--Cognitive Subscale totals (ADAS-Cog 11/13; ADASQ4 word recognition). Higher scores indicate worse impairment for CDR-SB, FAQ, TRABSCOR, and ADAS-Cog, whereas higher scores indicate better performance for MMSE, MoCA, and recall-based memory scores; RAVLT forgetting metrics reflect greater memory loss when higher. MRI scans and structured metadata were integrated to enhance the analysis.

\begin{table}[t]
\centering
\caption{Demographic, cognitive, and clinical characteristics of participants by AD progression group.}
\label{tab:tab1}
\begin{threeparttable}
\setlength{\aboverulesep}{-0.6pt}
\setlength{\belowrulesep}{0pt}
\resizebox{\textwidth}{!}{%
\begin{tabular}{l|ccccc}
\toprule
AD progression group & CN to CN & CN to MCI & MCI to MCI & MCI to AD & AD to AD \\
\midrule
Number of participants              & 160              & 27               & 434              & 117              & 106               \\
Number of scans                     & 755              & 43              & 1909             & 234              & 380              \\
Visits per participant             & 5.0 (2--6)   & 4.7 (2--6)    & 4.9 (1--9)    & 4.7 (2--8)    & 3.7 (2--7)    \\
Interval from baseline (month)      & 13.0 $\pm$ 15.0  & 20.6 $\pm$ 27.0  & 13.1 $\pm$ 15.4  & 17.6 $\pm$ 21.1  & 6.5 $\pm$ 10.0   \\
Age (year)                          & 72.7 $\pm$ 5.5   & 76.2 $\pm$ 7.0   & 71.0 $\pm$ 7.4   & 72.3 $\pm$ 7.3   & 75.4 $\pm$ 7.5   \\
Sex (female/male)                   & 61 / 55          & 10 / 15          & 119 / 158        & 41 / 66          & 37 / 57          \\
Education (year)                    & 16.3 $\pm$ 2.4   & 16.1 $\pm$ 2.6   & 16.1 $\pm$ 2.6   & 15.9 $\pm$ 3.0   & 15.7 $\pm$ 2.7   \\
Proportion of non-Hispanic/Latino   & 82.8\%           & 88.0\%           & 90.3\%           & 96.3\%           & 89.4\%           \\
\hdashline
CDR-SB                              & 0.1 $\pm$ 0.2    & 0.4 $\pm$ 0.6    & 1.2 $\pm$ 0.9    & 3.8 $\pm$ 2.5    & 5.3 $\pm$ 2.5    \\
MMSE                                & 28.9 $\pm$ 1.3   & 28.4 $\pm$ 1.7   & 28.1 $\pm$ 1.9   & 24.6 $\pm$ 4.0   & 21.9 $\pm$ 3.8   \\
MoCA                                & 26.2 $\pm$ 2.5   & 24.7 $\pm$ 2.9   & 24.2 $\pm$ 3.0   & 20.1 $\pm$ 4.3   & 16.6 $\pm$ 5.0   \\
FAQ                                 & 0.2 $\pm$ 0.8    & 0.6 $\pm$ 1.5    & 2.0 $\pm$ 3.2    & 10.3 $\pm$ 7.3   & 16.0 $\pm$ 7.3   \\
RAVLT Immediate                     & 46.4 $\pm$ 10.3  & 38.9 $\pm$ 10.6  & 38.4 $\pm$ 11.5  & 25.2 $\pm$ 8.3   & 21.5 $\pm$ 9.0   \\
RAVLT Learning                      & 5.8 $\pm$ 2.5    & 4.7 $\pm$ 2.7    & 4.9 $\pm$ 2.6    & 2.4 $\pm$ 2.1    & 1.7 $\pm$ 2.0    \\
RAVLT Forgetting                    & 3.6 $\pm$ 2.7    & 5.0 $\pm$ 2.4    & 4.5 $\pm$ 2.7    & 5.0 $\pm$ 2.1    & 4.2 $\pm$ 1.8    \\
RAVLT Percent Forgetting            & 34.6 $\pm$ 28.6  & 55.4 $\pm$ 28.8  & 53.0 $\pm$ 32.8  & 86.1 $\pm$ 24.0  & 91.8 $\pm$ 19.4  \\
LDELTOTAL                           & 14.3 $\pm$ 3.7   & 10.4 $\pm$ 3.7   & 9.2 $\pm$ 4.4    & 3.0 $\pm$ 3.9    & 1.6 $\pm$ 2.7    \\
TRABSCOR                            & 79.1 $\pm$ 40.7  & 91.2 $\pm$ 34.6  & 96.1 $\pm$ 52.8  & 155.9 $\pm$ 88.5 & 200.8 $\pm$ 88.0 \\
ADAS-Cog 11                         & 5.2 $\pm$ 2.8    & 7.2 $\pm$ 3.4    & 8.1 $\pm$ 4.3    & 16.8 $\pm$ 8.3   & 22.0 $\pm$ 8.9   \\
ADAS-Cog 13                         & 8.1 $\pm$ 4.3    & 11.9 $\pm$ 5.3   & 12.9 $\pm$ 6.5   & 25.8 $\pm$ 10.3  & 32.5 $\pm$ 10.7  \\
ADAS-Q4                             & 2.5 $\pm$ 1.8    & 4.0 $\pm$ 2.1    & 4.2 $\pm$ 2.4    & 7.7 $\pm$ 2.2    & 8.7 $\pm$ 1.8    \\
\bottomrule
\end{tabular}
}
\begin{tablenotes}[para,flushleft]
\footnotesize
Values are reported as mean $\pm$ standard deviation or mean (min--max).
\end{tablenotes}
\end{threeparttable}
\end{table}

\setlength{\textfloatsep}{11pt}

\subsection{Preprocessing}
\subsubsection{Image preprocessing}
We constructed paired image--text samples from longitudinal T1w MRI using a consistent spatial and intensity standardization pipeline. For spatial alignment, each 3D volume was registered to a fixed reference scan using Advanced Normalization Tools~\cite{avants2009ants} with a nonlinear SyN transformation, and the warped moving image was retained as the aligned volume. To further harmonize slice positioning across participants, we applied an additional translation along the superior--inferior ($z$) axis by adjusting the output affine so that the aligned volume origin matched the reference origin in $z$. All registered volumes were manually inspected for quality control (QC) prior to slice extraction; scans exhibiting severe artifacts/noise or substantial intensity inconsistency were excluded. From each QC-passed, $z$-adjusted volume, we extracted axial slices within a predefined index range ($z$-index 93--170, inclusive), yielding 78 slices per volume. We used all slice indices in this range and treated each slice as an individual sample. Each slice was converted to a single-channel grayscale image, intensity-normalized to $[0,1]$ using min--max scaling, and histogram matching (SimpleITK) was applied to match each normalized slice to a reference slice intensity distribution at the same slice index to reduce inter-scan brightness and contrast variability. All slices were then resized to a fixed in-plane matrix size of $256 \times 256$ to ensure a uniform input resolution. This preprocessing yielded 259{,}038 QC-retained two-dimensional (2D) slice images, each forming a slice--prompt pair with its associated text prompt.

\vspace{-0.78cm}

\subsubsection{Text preprocessing}
In parallel, we constructed a natural-language conditioning prompt for each slice using the diagnostic labels (CN, MCI, and AD) and the demographic, clinical, and cognitive attributes reported in Table~\ref{tab:tab1}. Each sample was indexed in a dataset manifest that links the processed slice to its corresponding conditioning prompt, enabling paired training. For each participant, we computed the interval (in months) from the first visit based on the MRI acquisition dates and used it as an interval attribute. During quality control, we resolved within-participant diagnostic inconsistencies by imposing a monotonic disease-severity constraint across visits (CN $<$ MCI $<$ AD). When a later visit exhibited an apparent diagnostic reversion (e.g., MCI$\rightarrow$CN), we reassigned that visit to the most recent prior diagnosis for the same participant that maintained non-decreasing severity over time.

\subsection{ADP-DiT architecture}

\setlength{\intextsep}{10pt}
\begin{figure}[h]
    \centering
    \includegraphics[width=1\textwidth]{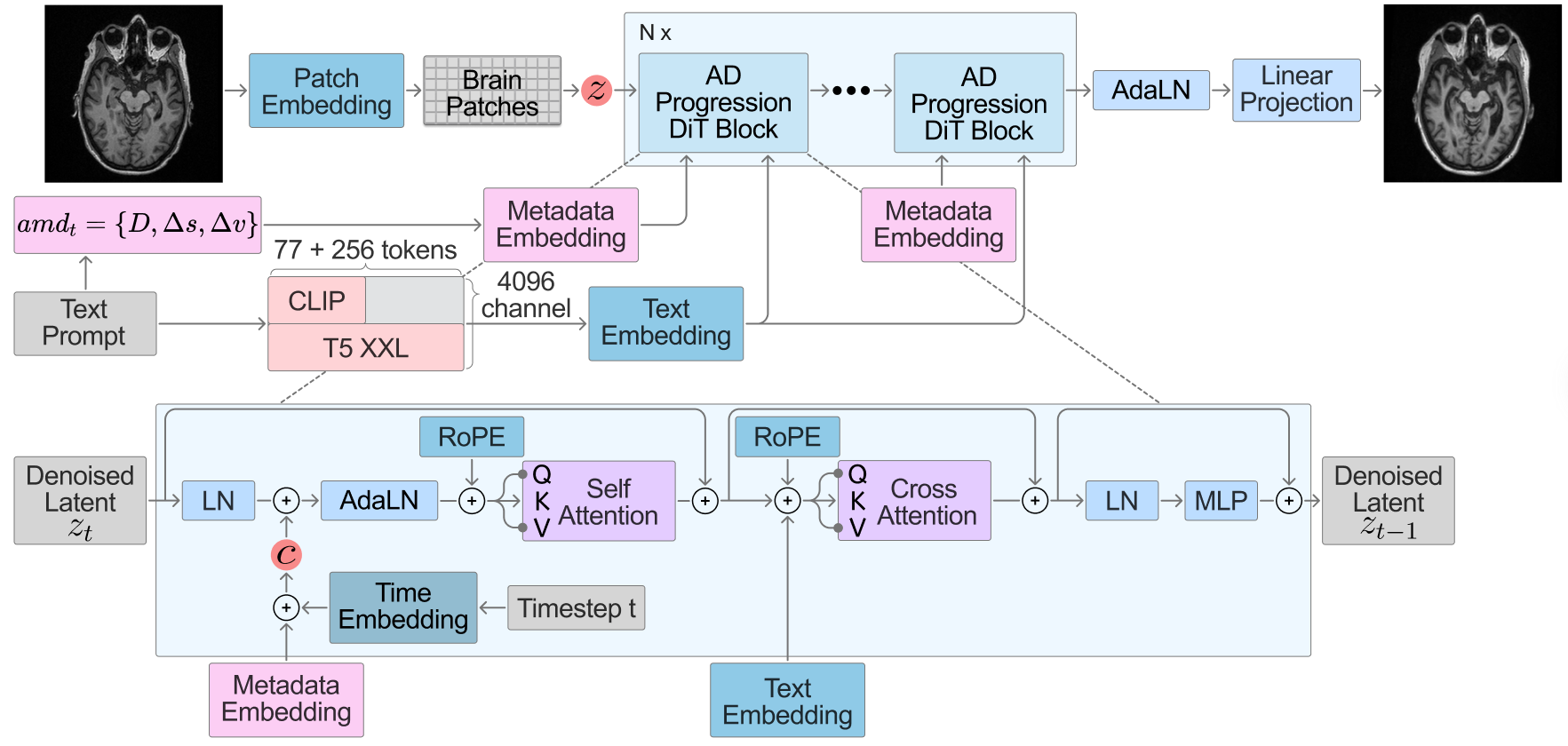}
    \caption{The overview of the proposed ADP-DiT framework. The model conditions on multiple text-encoders text embeddings and structured metadata to guide the denoising process in the latent space.}
    \label{fig:fig1}
\end{figure}

\subsubsection{Image encoding and latent representation}

To achieve high-fidelity synthesis while maintaining computational efficiency, we employ the pre-trained SDXL-VAE-FP16 Variational Autoencoder~\cite{podell2023sdxl}. This module serves as the interface between the high-dimensional pixel space and the computationally efficient latent space.

When an input brain MRI is provided, it is formatted as a 3-channel RGB image with a resolution of $256 \times 256$ pixels. The SDXL-VAE-FP16 encoder compresses this input into a low-dimensional latent representation $\mathbf{z} \in \mathbb{R}^{32 \times 32 \times 4}$, effectively reducing the spatial resolution by a factor of 8 while mapping the data into 4 latent channels. Following encoding, the latent vectors are multiplied by a scaling factor of 0.13025 to normalize the variance, ensuring stability during the diffusion process.

\subsubsection{Multimodal conditioning strategy}
Effective guidance of disease progression requires integrating diverse data modalities, including clinical text, longitudinal metadata, and demographic information. We propose a unified conditioning framework that fuses these elements into a coherent representation.

\vspace{-0.7cm}

\subsubsection{Text embedding}
Building on prior work showing that ensembling text encoders significantly improves generative performance~\cite{Balaji2022eDiff-I}, we employ a multi-text-encoder architecture to balance semantic understanding with medical precision. We combine OpenCLIP ViT-G/14~\cite{cherti2023reproducible} ($D_{\text{CLIP}} = 1280$, $L_{\text{CLIP}} = 77$) for vision-language semantic alignment and T5-XXL~\cite{raffel2020t5} ($D_{\text{T5}} = 4096$, $L_{\text{T5}} = 256$) for nuanced medical language understanding. The extended T5 context window accommodates complex descriptions, including disease stage and participant demographics, that are beyond the scope of conventional short-text prompts.

\vspace{-0.7cm}

\subsubsection{Temporal dynamics and metadata integration}
To capture the temporal progression of AD, we incorporate continuous time variables and clinical descriptors. Instead of training a separate auxiliary network, we employ a deterministic normalization function $f_\varphi$, implemented as Min-Max scaling, to directly map these longitudinal covariates into a model-compatible range of $[0, 1]$.

For a subject with multiple visits, we explicitly calculate the temporal and clinical transitions between the baseline visit $A_t$ and a follow-up visit $A_{t+1}$. We define the time interval $\Delta t$ as the duration in months between scans, and the clinical trajectory $\Delta s$ as the arithmetic difference in neuropsychological scores (i.e., $s_{t+1} - s_t$). Consequently, the Auxiliary Metadata (AMD) is formulated as a concatenated vector of the normalized static demographics $D$ and dynamic transitions: $amd_t = \{f_\varphi(D), f_\varphi(\Delta t), f_\varphi(\Delta s)\}$.

To integrate this metadata with the global context, we construct a unified embedding vector $v_{\text{meta}}$ by concatenating the pooled text embedding from the T5 encoder $y_{\text{pooled}}$, image resolution metadata $v_{\text{res}}$, and our normalized auxiliary metadata $amd_t$:
\vspace{-0.3cm}

\begin{equation}
    v_{\text{meta}} = \operatorname{Concat}(y_{\text{pooled}}, v_{\text{res}}, amd_t)
\end{equation}

This combined vector is processed by the Metadata Embedding module, which consists of a multi-layer perceptron (MLP) utilizing the sigmoid-weighted linear units (SiLU) activation function. The MLP projects the input into a high-dimensional space ($4 \times$ hidden size) before compressing it back to the model dimension, thereby ensuring rich feature interactions. Finally, this processed vector is added to the timestep embedding $t_{\text{emb}}$ to form the global conditioning vector $c_{\text{global}}$:

\vspace{-0.2cm}

\begin{equation}
    c_{\text{global}} = t_{\text{emb}} + \operatorname{MLP}(v_{\text{meta}})
\end{equation}

\vspace{0.1cm}

This mechanism ensures that the diffusion process is modulated not only by the noise level ($t$) but also by the semantic context and specific disease progression markers, such as the exact time elapsed since the baseline scan.

\subsection{ADP-DiT backbone architecture}
We adopt the DiT~\cite{peebles2023dit} as our generative backbone, modifying it to support multimodal conditioning and to enhance spatial modeling via a hybrid conditioning mechanism.

\vspace{-0.3cm}

\subsubsection{Transformer blocks with multi-head cross-attention}
Unlike standard DiT implementations that rely solely on Adaptive Layer Normalization (AdaLN) for class-conditional control, ADP-DiT integrates text and metadata via a dual-mechanism approach. Global conditioning information—encapsulated in the unified vector $c_{\text{global}}$—is modulated through AdaLN at the self-attention stage. For text conditioning, we first project the T5-XXL sequence embeddings ($H_{\text{T5}}$) into the CLIP embedding space via a two-layer MLP with SiLU activation, then concatenate them with CLIP tokens ($H_{\text{CLIP}}$) along the sequence dimension. The resulting unified text representation serves as keys and values for multi-head cross-attention:

\vspace{-0.2cm}

\begin{equation}
    H_{\text{fused}} = \operatorname{CrossAttn}\bigl(\operatorname{LN}(\mathbf{z}),\; \operatorname{Concat}(H_{\text{CLIP}}, \operatorname{MLP}(H_{\text{T5}}))\bigr)
\end{equation}

\vspace{0.1cm}

This design enables the attention mechanism to implicitly learn adaptive weighting between CLIP's semantic guidance and T5's linguistic precision based on the query content, while AdaLN maintains global context consistency.

\vspace{-0.4cm}

\subsubsection{RoPE for spatial alignment}
To enhance the capture of spatial dependencies within brain MRI data, we incorporate Rotary Positional Embeddings (RoPE)~\cite{su2024roformer}. Our empirical results indicate that applying RoPE explicitly to image feature maps yields superior structural fidelity. Within the self-attention mechanism, we apply RoPE to both the query ($\mathbf{Q}$) and key ($\mathbf{K}$) matrices, ensuring that relative position information is directly encoded into the attention scores:

\vspace{-0.2cm}

\begin{equation}
    \text{Attn}(\mathbf{Q}^{\text{RoPE}}, \mathbf{K}^{\text{RoPE}}, \mathbf{V}) = \text{softmax}\left(\frac{\mathbf{Q}^{\text{RoPE}} (\mathbf{K}^{\text{RoPE}})^T}{\sqrt{d}}\right) \mathbf{V}
\end{equation}

\vspace{0.1cm}

Furthermore, in the cross-attention module, we implement a selective application strategy where RoPE is applied solely to the image query embeddings ($\mathbf{Q}_{\text{img}}$). **This design reflects the modal difference: while MRI slices possess fixed 2D spatial geometry, text prompts are sequential and lack inherent spatial coordinates.** Therefore, the text keys and values ($\mathbf{K}_{\text{text}}, \mathbf{V}_{\text{text}}$) remain unrotated. By injecting positional information exclusively into the image queries, we allow each latent image patch to attend to the clinical context with explicit awareness of its anatomical location.

\subsection{Diffusion process and reconstruction}
The ADP-DiT model refines the latent representation $\mathbf{z}$ through iterative denoising steps. At each timestep, noise is removed under the guidance of the unified conditioning signals ($c_{\text{global}}$ and text context). During training, we adopt the v-prediction parameterization~\cite{salimans2022progressive} for improved stability, optimizing the model to predict velocity in the latent space. Finally, the denoised latent vector is decoded by the VAE decoder to reconstruct the high-resolution MRI, visualizing the expected disease progression. The code will be made available upon acceptance at \url{https://github.com/labhai/ADP-DiT}.

\section{Experiments}
\subsection{AD progression brain image generation} 
\subsubsection{Dataset construction and input formatting}
The input data consists of longitudinal 2D axial MRI slices preprocessed to a resolution of $256 \times 256$. To align with the input requirements of the pre-trained SDXL-VAE-FP16 encoder, the original grayscale images were converted into a 3-channel RGB format. Consequently, the encoder compresses these inputs into compact 4-channel latent representations ($\mathbf{z} \in \mathbb{R}^{32 \times 32 \times 4}$), which are subsequently scaled by a factor of 0.13025 to normalize variance for stable diffusion training.

For conditioning, each image is paired with a structured clinical metadata prompt that encapsulates demographics, diagnosis, visit timing, and thirteen specific neuropsychological scores—including CDR-SB, ADAS-13, and MMSE—enabling the model to learn fine-grained correlations between clinical metrics and anatomical features. Furthermore, to improve generalization and simulate real-world acquisition variability, we employed a robust augmentation pipeline comprising nine MRI-specific transformations, such as elastic deformation to model anatomical variance and bias field simulation for intensity inhomogeneities. The dataset was partitioned at the participant level into 80\% for training and 20\% for testing. To perform model selection and hyperparameter tuning, we initially held out a validation subset consisting of one representative participant for each disease progression scenario from the training partition. Following the identification of optimal hyperparameters, the final model was retrained using the complete training dataset—incorporating both the initial training and validation samples—to maximize data utilization. To prevent information leakage across longitudinal scans, all scans and associated metadata from the same participant were assigned to a single split. The training set contained 555 participants, comprising 2{,}569 longitudinal scans and 200{,}382 derived 2D slices. The test set contained 157 participants, comprising 752 scans and 58{,}656 2D slices.

\vspace{-0.65cm}

\subsubsection{Implementation details and experimental setup}
We implemented the DiT architecture~\cite{peebles2023dit} (approximately 1.9B parameters) with a depth of 40, hidden size of 1408, and 16 attention heads. Training was conducted for 22,686 steps on eight NVIDIA RTX 5880 Ada Generation GPUs, utilizing a global batch size of 1,024. We used the AdamW optimizer~\cite{loshchilov2017decoupled,kingma2014adam} ($lr=5 \times 10^{-5}$) combined with a cosine annealing with warm restarts scheduler, incorporating a 5\% linear warmup phase.

The model optimizes the weighted mean squared error (MSE) objective in the latent space:

\vspace{-0.2cm}

\begin{equation}
    \mathcal{L}(\theta) = \mathbb{E}_{t, \mathbf{z}_0, \epsilon} \left[ w(t) \cdot || \mathbf{y}_{\text{target}} - f_\theta(\mathbf{z}_t, t, c) ||^2 \right]
\end{equation}

\vspace{0.1cm}

where $\mathbf{z}_t$ represents the noisy latent state at timestep $t$, $\mathbf{z}_0$ is the clean latent representation encoded by the VAE, and $\epsilon \sim \mathcal{N}(0, \mathbf{I})$ is the Gaussian noise. The term $c$ denotes the unified multimodal conditioning context, and $w(t)$ is a time-dependent weighting function. Adopting velocity parameterization ($\mathbf{v}$-prediction)~\cite{salimans2022progressive}, the target $\mathbf{y}_{\text{target}}$ corresponds to the ground truth velocity $\mathbf{v}_t$. For image synthesis during inference, we used the DPM-Solver++ (2M) sampler~\cite{lu2025dpmsolver} with Karras noise scheduling, employing a classifier-free guidance scale of 4.5~\cite{ho2022classifierfree}.

For evaluation, we benchmarked ADP-DiT against competing models including SwinUNETR-V2~\cite{hatamizadeh2023swinunetr}, Stable Diffusion 2.1~\cite{rombach2022sd}, FCDiffusion~\cite{wu2024fcdiffusion}, VQGAN-CLIP~\cite{crowson2022vqgan}, and Diffusion-CLIP~\cite{kim2022diffusionclip}. We also report an identity baseline (Baseline), where the model output is identical to the input image. We kept the same structure, parameters, and environment as in the official code wherever possible. Exceptions include the use of the regression head and the MSE loss in segmentation-purpose models. All models were trained on an identical dataset partition and used optimized hyperparameters to ensure fair comparison. We rigorously evaluated the predictive performance of ADP-DiT against established baseline methods using standard image quality metrics: structural similarity index (SSIM), peak signal-to-noise ratio (PSNR), and MSE. We report SSIM as the primary metric to emphasize structural fidelity relevant to disease monitoring, with PSNR and MSE as secondary metrics.

\subsection{Results}

As summarized in Table~\ref{tab:tab2}, our ADP-DiT model achieves competitive performance across all metrics, with a SSIM of 0.8739, a PSNR of 29.32, and an MSE of 0.0024. This indicates that our model generates images with high structural fidelity and low noise levels, surpassing general-domain generative models such as Stable Diffusion~\cite{rombach2022sd} and VQGAN-CLIP~\cite{crowson2022vqgan}.
We further performed statistical significance tests for each metric to compare ADP-DiT with the competing models; the results are reported in Table~\ref{tab:tab2}. The tests show that ADP-DiT achieves statistically significant improvements over the DiT baseline and all text-guided generative models across SSIM, PSNR, and MSE.

To provide a deeper insight into disease-specific dynamics, Table~\ref{progression_tab} details the performance across specific disease progression scenarios. The model maintains high structural consistency in stable conditions (e.g., CN $\to$ CN and AD $\to$ AD), consistently achieving SSIM scores exceeding 0.86. In the most challenging progression scenario (MCI $\to$ AD), which involves significant anatomical atrophy and morphological alterations, the SSIM decreases to 0.8227. This deviation is expected given the complexity of predicting non-linear structural changes; however, the model effectively captures the transitional features, validating its sensitivity to disease-specific pathology rather than merely reconstructing static anatomy.

{
\setlength{\intextsep}{0pt}
\begin{table}[H]
\centering
\caption{Quantitative comparison of ADP-DiT against state-of-the-art models. \textbf{Bold} indicates the best performance among generative models, and \underline{underline} indicates the second best. Models below the dashed line are sorted by SSIM.}
\label{tab:tab2}
\begin{threeparttable}
\setlength{\aboverulesep}{-1pt}
\setlength{\belowrulesep}{0pt}
\begin{tabularx}{\textwidth}{p{3.5cm}| >{\centering\arraybackslash}X >{\centering\arraybackslash}X >{\centering\arraybackslash}X}
\toprule
Models & SSIM$\uparrow$ & PSNR$\uparrow$ & MSE$\downarrow$ \\
\midrule
Baseline (Out = In)                         & \underline{0.8739} $\pm$ 0.0640 \;\:          & 26.14 $\pm$ 2.654$^{**}$          & 0.0030 $\pm$ 0.0024$^{**}$ \\
Diffusion Transformer~\cite{peebles2023dit}  & 0.7652 $\pm$ 0.0599$^{**}$          & 23.24 $\pm$ 1.840$^{**}$          & 0.0052 $\pm$ 0.0024$^{**}$ \\
SwinUNETR-V2~\cite{hatamizadeh2023swinunetr} & 0.8535 $\pm$ 0.0686$^{**}$ & \underline{28.37} $\pm$ 3.108$^{**}$ & \textbf{0.0019} $\pm$ 0.0019 \;\: \\
\hdashline
FCDiffusion~\cite{wu2024fcdiffusion}         & 0.5474 $\pm$ 0.1394$^{**}$          & 20.55 $\pm$ 2.562$^{**}$          & 0.0106 $\pm$ 0.0080$^{**}$ \\
Diffusion-CLIP~\cite{kim2022diffusionclip}   & 0.6556 $\pm$ 0.1124$^{**}$          & 24.33 $\pm$ 3.707$^{**}$          & 0.0051 $\pm$ 0.0041$^{**}$ \\
VQGAN-CLIP~\cite{crowson2022vqgan}           & 0.7463 $\pm$ 0.0594$^{**}$          & 22.97 $\pm$ 1.816$^{**}$          & 0.0055 $\pm$ 0.0024$^{**}$ \\
Stable Diffusion 2.1~\cite{rombach2022sd}    & 0.7952 $\pm$ 0.0799$^{**}$          & 23.44 $\pm$ 2.649$^{**}$          & 0.0055 $\pm$ 0.0046$^{**}$ \\
\textbf{ADP\mbox{-}DiT (Ours)}               & \textbf{0.8739} $\pm$ 0.0761 \;\:   & \textbf{29.32} $\pm$ 6.811 \;\:   & \underline{0.0024} $\pm$ 0.0024 \;\: \\
\bottomrule
\end{tabularx}
\begin{tablenotes}[para,flushleft]
\footnotesize
Values are presented as mean $\pm$ standard deviation on the test set. $p$\mbox{-}values were calculated to compare each model with ADP\mbox{-}DiT using paired one-sided $t$-tests over matched samples. \\Asterisks ($^*$) denote statistical significance. $^{*}$ denotes $p<0.05$, and $^{**}$ denotes $p<0.005$.
\end{tablenotes}
\end{threeparttable}
\end{table}
\setlength{\intextsep}{0pt}
}

\vspace{-0.3cm}

\begin{table}[H]
\centering
\caption{Performance of ADP-DiT across progression groups.}
\label{progression_tab}
\begin{threeparttable}
\setlength{\tabcolsep}{3pt}
\renewcommand{\arraystretch}{1.2}
\setlength{\aboverulesep}{-0.6pt}
\setlength{\belowrulesep}{0pt}

\begin{tabularx}{\textwidth}{
  >{\centering\arraybackslash}p{1.7cm}
  | >{\centering\arraybackslash}p{1.8cm}
  | >{\centering\arraybackslash}p{0.5cm}
    >{\centering\arraybackslash}p{0.6cm}
  | >{\centering\arraybackslash}p{2.1cm}        
    >{\centering\arraybackslash}p{1.9cm}        
    >{\centering\arraybackslash}p{2.1cm}        
@{}}
\toprule
Progression & Interval (Mo.) & N & Scan & SSIM$\uparrow$ & PSNR$\uparrow$ & MSE$\downarrow$ \\
\hline
CN $\to$ CN   & 8.50 $\pm$ 7.02  & 30 & 171 & 0.8695 $\pm$ 0.0804 & 29.09 $\pm$ 6.739 & 0.0025 $\pm$ 0.0029 \\
CN $\to$ MCI  & 13.31 $\pm$ 9.19 & 5 & 10  & 0.8131 $\pm$ 0.0233 & 24.58 $\pm$ 1.256 & 0.0036 $\pm$ 0.0010 \\
MCI $\to$ MCI & 7.49 $\pm$ 5.59  & 81  & 432  & 0.8779 $\pm$ 0.0727 & 29.60 $\pm$ 6.850 & 0.0023 $\pm$ 0.0021 \\
MCI $\to$ AD  & 11.08 $\pm$ 6.60 & 21 & 53 & 0.8227 $\pm$ 0.0685 & 24.73 $\pm$ 2.697 & 0.0041 $\pm$ 0.0034 \\
AD $\to$ AD   & 7.43 $\pm$ 6.09  & 20 & 86 & 0.8895 $\pm$ 0.0761 & 30.71 $\pm$ 7.236 & 0.0020 $\pm$ 0.0018 \\
\bottomrule
\end{tabularx}

\begin{tablenotes}[para,flushleft]
\small
\item Values are presented as mean $\pm$ standard deviation on test set.
\end{tablenotes}
\end{threeparttable}
\end{table}
\setlength{\intextsep}{5pt}

\vspace{-0.1cm}

As shown in Table~\ref{tab:tab4}, we further analyze performance by stratifying test pairs according to the follow-up interval $\Delta t$. Overall, all methods exhibit a monotonic degradation as $\Delta t$ increases, reflecting the growing difficulty of modeling longer-term, non-linear anatomical changes. In the shortest interval bin ($0 \le \Delta t < 12$), ADP-DiT achieves the highest SSIM (0.8975) and PSNR (31.52) while maintaining low MSE (0.0018), outperforming all text-guided generative baselines in this regime. For longer intervals ($12 \le \Delta t < 24$, $24 \le \Delta t < 36$, and $36 \le \Delta t$), ADP-DiT consistently remains the strongest generative model in SSIM, sustaining despite increasing temporal gaps. These results indicate that ADP-DiT preserves robust interval-conditioned generation quality across a wide range of follow-up durations, while providing the controllability and diversity absent in purely reconstructive approaches.

\vspace{-0.1cm}

\section{Discussion}
These empirical results demonstrate the feasibility and efficacy of using a transformer-based diffusion model to synthesize AD progression conditioned on clinical metadata. Unlike traditional registration-based deformation models or GANs, which often struggle with long-range dependencies and multi-modal integration, our ADP-DiT framework successfully bridges the gap between discrete clinical data and high-dimensional neuroimaging. We provide evidence that integrating comprehensive clinical descriptions—ranging from demographics to granular cognitive scores—directly into the generative process enables the synthesis of biologically plausible changes in brain morphology over time.

\setlength{\intextsep}{0.1cm}

\begin{table}[!thbp]
\centering
\setlength{\aboverulesep}{-0.6pt}
\setlength{\belowrulesep}{0pt}
\caption{Benchmark results comparing ADP-DiT with state-of-the-art methods across specific time intervals. \textbf{Bold} indicates the best performance among the compared methods, and \underline{underline} indicates the second best. Here, $\Delta t$ denotes the time elapsed (in months) between the baseline input and the target follow-up scan.}
\label{tab:tab4}
\footnotesize
\setlength{\tabcolsep}{2.5pt}
\renewcommand{\arraystretch}{1.1}
\begin{tabularx}{\textwidth}{
  >{\centering\arraybackslash}p{1.8cm}
  | >{\raggedright\arraybackslash}p{3.3cm}
  | >{\centering\arraybackslash}p{2.2cm}
    >{\centering\arraybackslash}p{1.9cm}
    >{\centering\arraybackslash}p{2.2cm}
}
\toprule
Interval (Mo.) & Models & SSIM$\uparrow$ & PSNR$\uparrow$ & MSE$\downarrow$ \\
\midrule

\multirow{7}{*}{\shortstack{$0 \le \Delta t < 12$ \\ \footnotesize
\\
\\
\#subj: 151 \\
\#slice: 35334
}}
& Diffusion Transformer~\cite{peebles2023dit}  & 0.7748 $\pm$ 0.0532 & 23.51 $\pm$ 1.788 & 0.0048 $\pm$ 0.0021 \\
& SwinUNETR-V2~\cite{hatamizadeh2023swinunetr} & \underline{0.8663} $\pm$ 0.0632 & \underline{29.29} $\pm$ 2.954 & \textbf{0.0015} $\pm$ 0.0014 \\
& FCDiffusion~\cite{wu2024fcdiffusion}         & 0.5634 $\pm$ 0.1392 & 20.82 $\pm$ 2.578 & 0.0100 $\pm$ 0.0076 \\
& Diffusion-CLIP~\cite{kim2022diffusionclip}   & 0.6651 $\pm$ 0.1168 & 25.14 $\pm$ 3.973 & 0.0044 $\pm$ 0.0037 \\
& VQGAN-CLIP~\cite{crowson2022vqgan}           & 0.7553 $\pm$ 0.0523 & 23.21 $\pm$ 1.774 & 0.0052 $\pm$ 0.0021 \\
& Stable Diffusion 2.1~\cite{rombach2022sd}    & 0.7172 $\pm$ 0.1151 & 24.41 $\pm$ 2.706 & 0.0044 $\pm$ 0.0033 \\
& ADP-DiT (Ours)                               & \textbf{0.8975} $\pm$ 0.0733 & \textbf{31.52} $\pm$ 7.529 & \underline{0.0018} $\pm$ 0.0021 \\
\midrule

\multirow{7}{*}{\shortstack{$12 \le \Delta t < 24$ \\ \footnotesize
\\
\\
\#subj: 103 \\
\#slice: 10062
}}
& Diffusion Transformer~\cite{peebles2023dit}  & 0.7533 $\pm$ 0.0651 & 22.96 $\pm$ 1.736 & 0.0055 $\pm$ 0.0026 \\
& SwinUNETR-V2~\cite{hatamizadeh2023swinunetr} & \underline{0.8398} $\pm$ 0.0677 & \textbf{27.46} $\pm$ 2.632 & \textbf{0.0022} $\pm$ 0.0021 \\
& FCDiffusion~\cite{wu2024fcdiffusion}         & 0.5239 $\pm$ 0.1360 & 20.11 $\pm$ 2.481 & 0.0117 $\pm$ 0.0086 \\
& Diffusion-CLIP~\cite{kim2022diffusionclip}   & 0.6493 $\pm$ 0.1039 & 23.23 $\pm$ 2.730 & 0.0058 $\pm$ 0.0042 \\
& VQGAN-CLIP~\cite{crowson2022vqgan}           & 0.7341 $\pm$ 0.0655 & 22.72 $\pm$ 1.744 & 0.0058 $\pm$ 0.0026 \\
& Stable Diffusion 2.1~\cite{rombach2022sd}    & 0.6607 $\pm$ 0.1332 & 22.59 $\pm$ 2.364 & 0.0064 $\pm$ 0.0045 \\
& ADP-DiT (Ours)                               & \textbf{0.8437} $\pm$ 0.0641 & \underline{26.28} $\pm$ 3.266 & \underline{0.0030} $\pm$ 0.0026 \\
\midrule

\multirow{7}{*}{\shortstack{$24 \le \Delta t < 36$ \\ \footnotesize
\\
\\
\#subj: 74 \\
\#slice: 5928
}}
& Diffusion Transformer~\cite{peebles2023dit}  & 0.7485 $\pm$ 0.0634 & 22.63 $\pm$ 1.862 & 0.0060 $\pm$ 0.0031 \\
& SwinUNETR-V2~\cite{hatamizadeh2023swinunetr} & \underline{0.8348} $\pm$ 0.0743 & \textbf{26.82} $\pm$ 2.917 & \textbf{0.0027} $\pm$ 0.0027 \\
& FCDiffusion~\cite{wu2024fcdiffusion}         & 0.5215 $\pm$ 0.1356 & 20.07 $\pm$ 2.439 & 0.0117 $\pm$ 0.0085 \\
& Diffusion-CLIP~\cite{kim2022diffusionclip}   & 0.6426 $\pm$ 0.0952 & 23.12 $\pm$ 2.777 & 0.0061 $\pm$ 0.0045 \\
& VQGAN-CLIP~\cite{crowson2022vqgan}           & 0.7303 $\pm$ 0.0644 & 22.43 $\pm$ 1.815 & 0.0062 $\pm$ 0.0030 \\
& Stable Diffusion 2.1~\cite{rombach2022sd}    & 0.6427 $\pm$ 0.1332 & 22.11 $\pm$ 2.532 & 0.0074 $\pm$ 0.0056 \\
& ADP-DiT (Ours)                               & \textbf{0.8380} $\pm$ 0.0607 & \underline{25.68} $\pm$ 3.281 & \underline{0.0035} $\pm$ 0.0031 \\
\midrule

\multirow{7}{*}{\shortstack{$\Delta t \ge 36$ \\ \footnotesize 
\\
\\
\#subj: 73 \\
\#slice: 7332
}}
& Diffusion Transformer~\cite{peebles2023dit}  & 0.7485 $\pm$ 0.0698 & 22.76 $\pm$ 1.943 & 0.0058 $\pm$ 0.0026 \\
& SwinUNETR-V2~\cite{hatamizadeh2023swinunetr} & \underline{0.8256} $\pm$ 0.0745 & \textbf{26.44} $\pm$ 2.819 & \textbf{0.0028} $\pm$ 0.0023 \\
& FCDiffusion~\cite{wu2024fcdiffusion}         & 0.5227 $\pm$ 0.1364 & 20.23 $\pm$ 2.513 & 0.0113 $\pm$ 0.0081 \\
& Diffusion-CLIP~\cite{kim2022diffusionclip}   & 0.6287 $\pm$ 0.1086 & 22.87 $\pm$ 3.025 & 0.0065 $\pm$ 0.0045 \\
& VQGAN-CLIP~\cite{crowson2022vqgan}           & 0.7327 $\pm$ 0.0694 & 22.60 $\pm$ 1.910 & 0.0060 $\pm$ 0.0025 \\
& Stable Diffusion 2.1~\cite{rombach2022sd}    & 0.6022 $\pm$ 0.1506 & 21.51 $\pm$ 2.544 & 0.0085 $\pm$ 0.0063 \\
& ADP-DiT (Ours)                               & \textbf{0.8305} $\pm$ 0.0706 & \underline{25.75} $\pm$ 3.818 & \underline{0.0035} $\pm$ 0.0024 \\
\bottomrule
\end{tabularx}

\begin{tablenotes}[para,flushleft]
\small
\item Values are presented as mean $\pm$ standard deviation on test set. 
\end{tablenotes}
\end{table}

\vspace{0.3cm}

\begin{figure}[H]
    \centering
    \includegraphics[width=\textwidth]{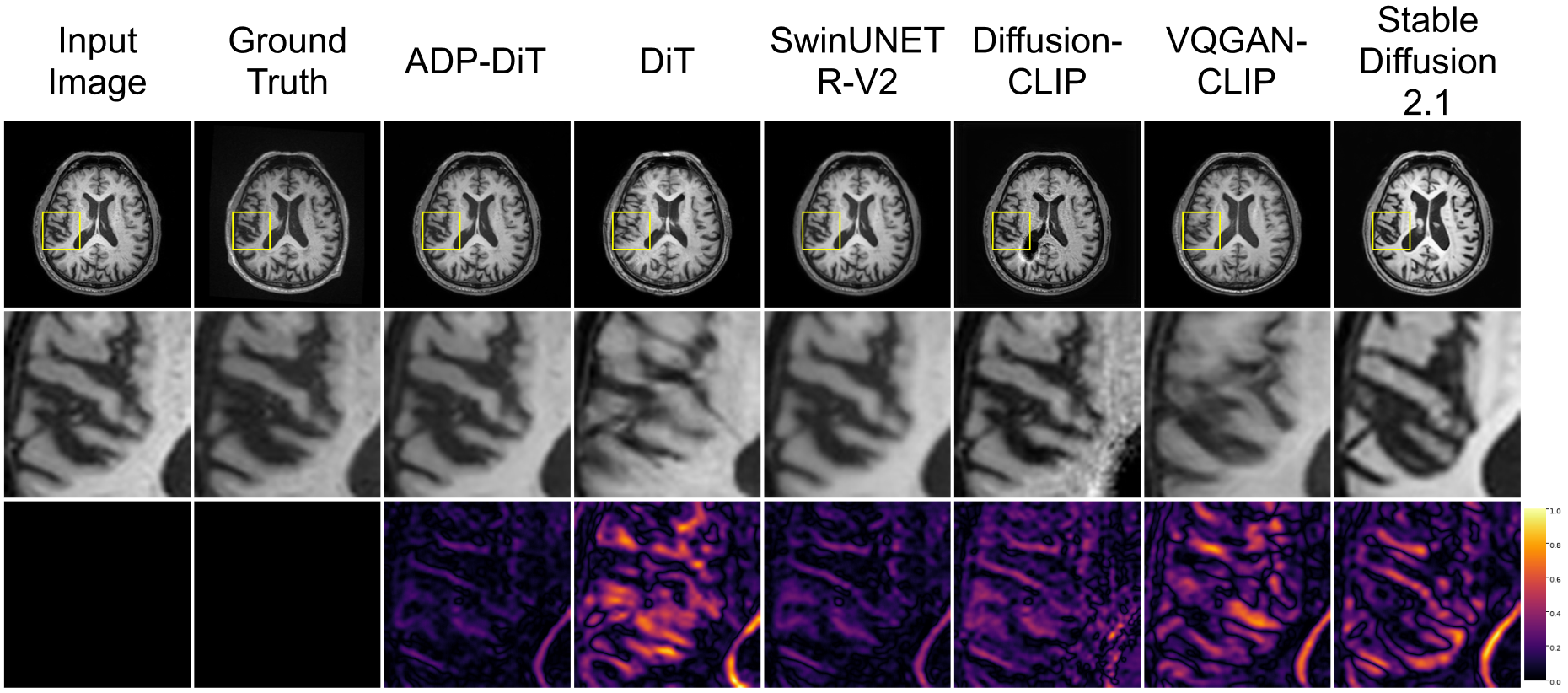}
    \caption{Visual comparison of ADP-DiT against benchmark models. The absolute error maps highlight the voxel-wise progression predicted by each model relative to the ground truth, demonstrating the superior accuracy of ADP-DiT.}
    \label{fig:fig2}
\end{figure}

\begin{figure}[!t]
    \centering
    \includegraphics[width=0.9\textwidth]{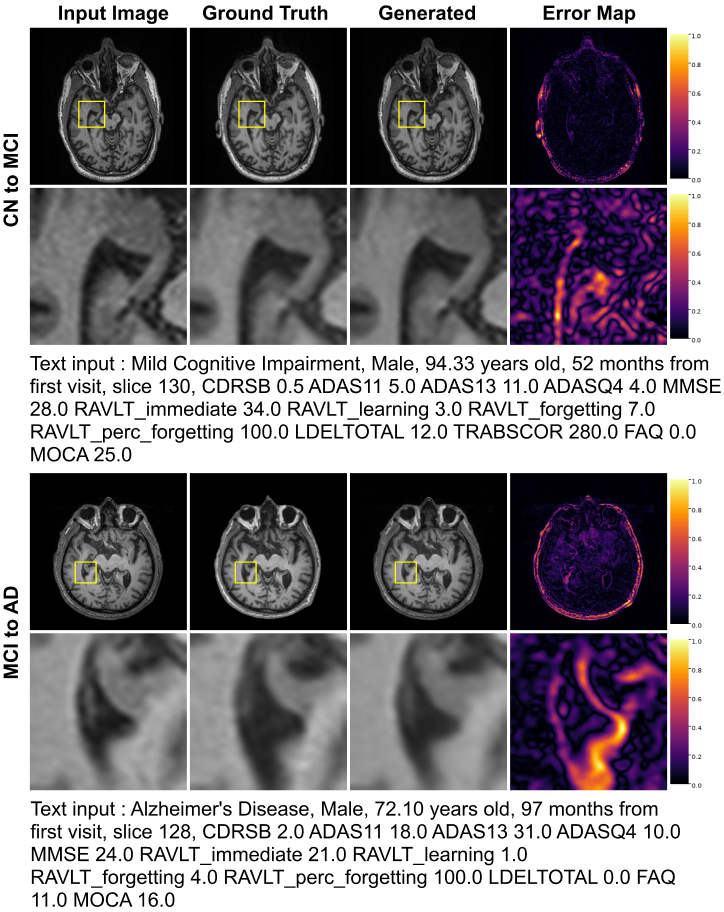}
    \caption{Results of ADP-DiT for AD progression. The absolute error map visualizes voxel-wise discrepancies between the generated output and the ground truth.}
    \label{fig:fig3}
\end{figure}

As detailed in Table~\ref{tab:tab4}, SSIM shows an overall decreasing trend as the follow-up interval $\Delta t$ increases, reflecting the growing difficulty of synthesizing longer-horizon anatomical change. PSNR and MSE are highly sensitive to small spatial misalignments and do not necessarily reflect anatomical plausibility; accordingly, conservative predictors such as SwinUNETR-V2 achieve lower MSE and higher PSNR in the longer-interval bins (Table~\ref{tab:tab4}). We therefore treat SSIM as the primary metric and report PSNR/MSE as complementary measures, prioritizing structural fidelity under increasing temporal gaps.\\
\indent Complementing these quantitative metrics, Fig.~\ref{fig:fig2} presents representative qualitative comparisons and corresponding absolute error maps. The absolute error maps reveal that ADP-DiT localizes residuals primarily around progression-relevant anatomical boundaries, including ventricular enlargement and cortical atrophy—hallmarks of AD progression—rather than introducing diffuse global artifacts. In our qualitative inspection, SwinUNETR-V2~\cite{hatamizadeh2023swinunetr} frequently produces follow-ups that remain close to the input anatomy, which can yield low MSE while underexpressing progression-related change; this behavior is consistent with the strong performance of the input-preserving model in pixel-wise metrics. By contrast, ADP-DiT more faithfully reflects AD-related progression effects, particularly around the lateral ventricles, aligning the generated anatomy with the target trajectory rather than merely preserving the baseline appearance.\\
\indent Fig.~\ref{fig:fig2} also highlights characteristic failure modes of competing text-guided generative models. FCDiffusion~\cite{wu2024fcdiffusion}, Stable Diffusion 2.1~\cite{rombach2022sd}, and Diffusion-CLIP~\cite{kim2022diffusionclip} often exhibit excessive ventricular expansion, distorted ventricular morphology, or peripheral noise, suggesting weaker spatial control over where and how anatomical changes should occur. Furthermore, the DiT~\cite{peebles2023dit} baseline can produce inconsistent progression trends (e.g., apparent lateral ventricle contraction in some examples), indicating that incorporating participant metadata and follow-up interval information is critical for consistent AD progression synthesis. By comparison, ADP-DiT preserves fine-grained structural details and shows more spatially coherent, anatomically constrained changes, which is also consistent with the role of Rotary Positional Embeddings (RoPE)~\cite{su2024roformer} in regulating spatial alignment and the dual-encoder conditioning (OpenCLIP + T5)~\cite{cherti2023reproducible,raffel2020t5} in improving clinical-text interpretability and control. Fig.~\ref{fig:fig3} further supports this point by showing that the remaining errors of ADP-DiT concentrate on subtle progression boundaries rather than broad structural corruption.

\vspace{-0.2cm}

\setlength{\textfloatsep}{0.3cm}

Fig.~\ref{fig:fig3} provides case-level evidence that ADP-DiT remains sensitive to progression-relevant anatomy even in diagnostically converting trajectories, which are typically more heterogeneous than diagnosis-stable groups. In the CN$\rightarrow$MCI example, an AD-related progression effect is reflected by enlargement of the temporal horn of the lateral ventricle; the corresponding absolute error map shows near-zero residuals within the expanded region, indicating that ADP-DiT focuses its modifications on the clinically meaningful change rather than introducing diffuse errors. In the MCI$\rightarrow$AD example, the target follow-up exhibits enlargement of the occipital horn of the lateral ventricle. However, the generated image is not identical to the target, ADP-DiT still introduces the correct directional change relative to the input baseline by producing an expanded occipital horn. Taken together, these qualitative observations suggest that—even when diagnosis-changing progression groups yield lower quantitative scores than diagnosis-stable groups—ADP-DiT can still generate images that explicitly encode salient effects of AD progression.\\
\indent The superior performance of ADP-DiT can be attributed to two key architectural choices grounded in the characteristics of medical data. First, the DiT backbone enables the capture of complex spatial hierarchies and global dependencies inherent in brain anatomy, which are often limited in CNN-based U-Net architectures due to their restricted receptive fields. Second, our multi-text-encoder strategy underscores the need for specialized semantic processing. While CLIP provides robust general visual alignment, it lacks the granularity to interpret specific medical metrics (e.g., CDR-SB scores). By augmenting it with T5-XXL, we effectively inject rich, high-context clinical information into the latent space. This synergy ensures that the generated images are not only visually realistic but also clinically coherent with the participant's metadata.\\
\indent Despite these achievements, our current approach entails certain limitations primarily stemming from computational constraints and intrinsic data characteristics. First, the model operates on 2D axial slices rather than complete 3D volumes. While 3D modeling is ideal for volumetric analysis, the computational cost of the self-attention mechanism in the Diffusion Transformer scales quadratically with sequence length, requiring VRAM capacities that exceed current standard hardware capabilities. Moreover, 2D slice training enables a larger and more diverse set of adequate batch sizes, which is crucial for stabilizing diffusion training. Second, the dataset exhibits inherent class imbalance and an irregular distribution of follow-up intervals. Specifically, clinical data is heavily skewed towards shorter intervals (e.g., 6--12 months), resulting in a scarcity of samples for long-term follow-ups. This data sparsity likely contributes to the performance degradation observed in longer intervals (as noted in Table~\ref{tab:tab4}), as the model has fewer opportunities to learn the complex, accumulated deformations associated with extended timeframes. Although our conditioning strategy mitigates this by embedding time intervals and disease states, the model's robustness on these underrepresented long-term transitions remains a challenge.\\
\indent To extend the clinical utility of ADP-DiT, future work will focus on several key directions. To address the temporal data imbalance, we plan to implement an interval-aware sampling strategy that ensures uniform data loading across different time bins during training. Furthermore, we intend to construct validation and test sets with stratified interval distributions to provide a more rigorous and unbiased assessment of long-term forecasting capabilities.\\
\indent We also aim to incorporate a downstream classification module to quantify the diagnostic value of the synthesized images and integrate multi-modal data, including PET scans or genetic markers such as APOE $\varepsilon4$, to enhance the predictive precision of disease trajectories. Finally, validating the model on external multi-cohort datasets will be essential to ensure generalizability across different scanner protocols and subject populations.

\section{Conclusion}
We introduced ADP-DiT, a text-conditioned Diffusion Transformer for interval-controlled longitudinal brain MR image synthesis in Alzheimer’s disease progression.
It encodes diagnosis, months-from-baseline, and multi-domain neuropsychological scores as a natural-language prompt, enabling time-specific conditioning for follow-up synthesis.
Architecturally, ADP-DiT combines dual-encoder text conditioning with a hybrid injection strategy—cross-attention for fine-grained semantic guidance and adaptive layer normalization for global modulation—and incorporates rotary positional embeddings on image tokens to improve spatial consistency. Diffusion is performed in a pretrained SDXL-VAE latent space to support efficient, high-resolution reconstruction.
Across benchmarks, achieves strong image-fidelity performance and improves over a DiT baseline, while qualitatively reflecting progression-relevant anatomical changes (e.g., ventricular enlargement and hippocampus shrinking) under the provided prompts. 
Overall, the results suggest that integrating subject-specific longitudinal and clinical information via text conditioning, together with transformer-based latent diffusion, is a practical direction for longitudinal AD MRI synthesis and for generating follow-up images aligned with progression trajectories.

\section*{Acknowledgments} 
This work was supported by the National Institute of Health research projects (2024ER04\\0700 \& 2025ER040300); the National Supercomputing Center with supercomputing resources, including technical support (KSC-2024-CRE-0021 \& KSC-2025-CRE-0065); the High-Performance Computing Support project (RQT-25-070083), funded by the Ministry of Science and ICT, Republic of Korea; and Hankuk University of Foreign Studies Research Fund of 2025. Data used in preparation of this article were obtained from the Alzheimer’s Disease Neuroimaging Initiative (ADNI) database (adni.loni.usc.edu). As such, the investigators within the ADNI contributed to the design and implementation of ADNI and/or provided data, but did not participate in the analysis or writing of this report. A complete listing of ADNI investigators can be found at: \url{http://adni.loni.usc.edu/wp-content/uploads/how_to_apply/ADNI_Acknowledgement_List.pdf}

\end{document}